\documentclass[letterpaper, 10 pt, conference]{cls/ieeeconf}
\IEEEoverridecommandlockouts
\let\labelindent\relax

\usepackage[table,usenames,dvipsnames]{xcolor}      
\usepackage{enumitem}                               
\usepackage{extarrows}                              
\usepackage[noadjust]{cite}                         

\usepackage{amsmath,amssymb,amsfonts,amsthm,dsfont} 
\usepackage{algorithm,algorithmicx,listings}        
\usepackage[noend]{algpseudocode}			        

\usepackage{graphicx,tabularx,subcaption}
\usepackage[export]{adjustbox}
\usepackage{multirow,multicol,array,rotating,diagbox}

\usepackage[font={small}]{caption}
\usepackage[titletoc,title]{appendix}

\usepackage{comment}
\usepackage{balance}
\usepackage[breaklinks=true, colorlinks, bookmarks=true, citecolor=Black, urlcolor=Violet,linkcolor=Black]{hyperref}

\def\argmin{\mathop{\arg\min}\limits}

\newcommand{\indicator}{\mathds{1}}

\newcommand{\scaleMathLine}[2][1]{\resizebox{#1\linewidth}{!}{$\displaystyle{#2}$}}
\newcommand{\prl}[1]{\left(#1\right)}

\newcommand{\crl}[1]{\left\{#1\right\}}

\theoremstyle{definition}

\newtheorem*{assumption*}{Assumption}
\newtheorem*{problem*}{Problem}
\newtheorem{problem}{Problem}
\theoremstyle{remark}

\newtheorem*{solution*}{Solution}

\newcommand{\calC}{{\cal C}}

\newcommand{\calY}{{\cal Y}}

\newcommand{\bff}{\mathbf{f}}
\newcommand{\bfg}{\mathbf{g}}

\newcommand{\bfp}{\mathbf{p}}

\newcommand{\bfx}{\mathbf{x}}
\newcommand{\bfy}{\mathbf{y}}

\newcommand{\bfF}{\mathbf{F}}

\newcommand{\bfN}{\mathbf{N}}

\newcommand{\bfP}{\mathbf{P}}

\newcommand{\bfR}{\mathbf{R}}

\newcommand{\bfX}{\mathbf{X}}
\newcommand{\bfY}{\mathbf{Y}}
\newcommand{\bfZ}{\mathbf{Z}}

\newcommand{\bbN}{\mathbb{N}}

\newcommand{\bbR}{\mathbb{R}}

\def\thetitle{CORSAIR: Convolutional Object Retrieval and Symmetry-AIded Registration}
\def\theauthor{Tianyu Zhao, Qiaojun Feng, Sai Jadhav, Nikolay Atanasov}
\def\thekeywords{}

\hypersetup{
  pdfauthor={\theauthor},%
  pdftitle={\thetitle},%
  pdfsubject={\thetitle},%
  pdfkeywords={\thekeywords}
}

\IEEEoverridecommandlockouts

\title{\LARGE \bf \thetitle}

\author{Tianyu Zhao \and Qiaojun Feng \and Sai Jadhav
\and Nikolay Atanasov
\thanks{We gratefully acknowledge support from ARL DCIST CRA W911NF-17-2-0181 and ONR SAI N00014-18-1-2828.}%
\thanks{The authors are with the Department of Electrical and Computer Engineering, University of California San Diego, La Jolla, CA 92093, USA {\tt\small \{tiz007,qjfeng,smjadhav,natanasov\}@ucsd.edu}.}
}

\begin{document}
\maketitle

\begin{abstract}
This paper considers online object-level mapping using partial point-cloud observations obtained online in an unknown environment. We develop an approach for fully Convolutional Object Retrieval and Symmetry-AIded Registration (CORSAIR). Our model extends the Fully Convolutional Geometric Features model to learn a global object-shape embedding in addition to local point-wise features from the point-cloud observations. The global feature is used to retrieve a similar object from a category database, and the local features are used for robust pose registration between the observed and the retrieved object. Our formulation also leverages symmetries, present in the object shapes, to obtain promising local-feature pairs from different symmetry classes for matching. We present results from synthetic and real-world datasets with different object categories to verify the robustness of our method.
\end{abstract}

\section{Introduction}
\label{sec:introduction}

Advances in learning-based computer vision algorithms have enabled detection, classification, and segmentation of objects in images and videos with impressive accuracy. However, spatial and temporal perception of 3D environments at an object level using streaming sensory data remains a challenging task. It involves problems such as joint object pose and shape estimation from multi-view observations, data association of object instances across time and space, and compressed object shape representation for large-scale mapping. While general object reconstructions \cite{Runz_MaskFusion_2018} without a prior object model is affected by occlusions, sensor noise, or segmentation algorithm misclassification, utilizing CAD object models to fit the noise observations may enable efficient and accurate object-level maps \cite{Bowman_SemSLAM_2017,Avetisyan_Scan2CAD_2019}.

\begin{figure}[t]
  \includegraphics[width=\linewidth,trim=1 10 1 10,clip]{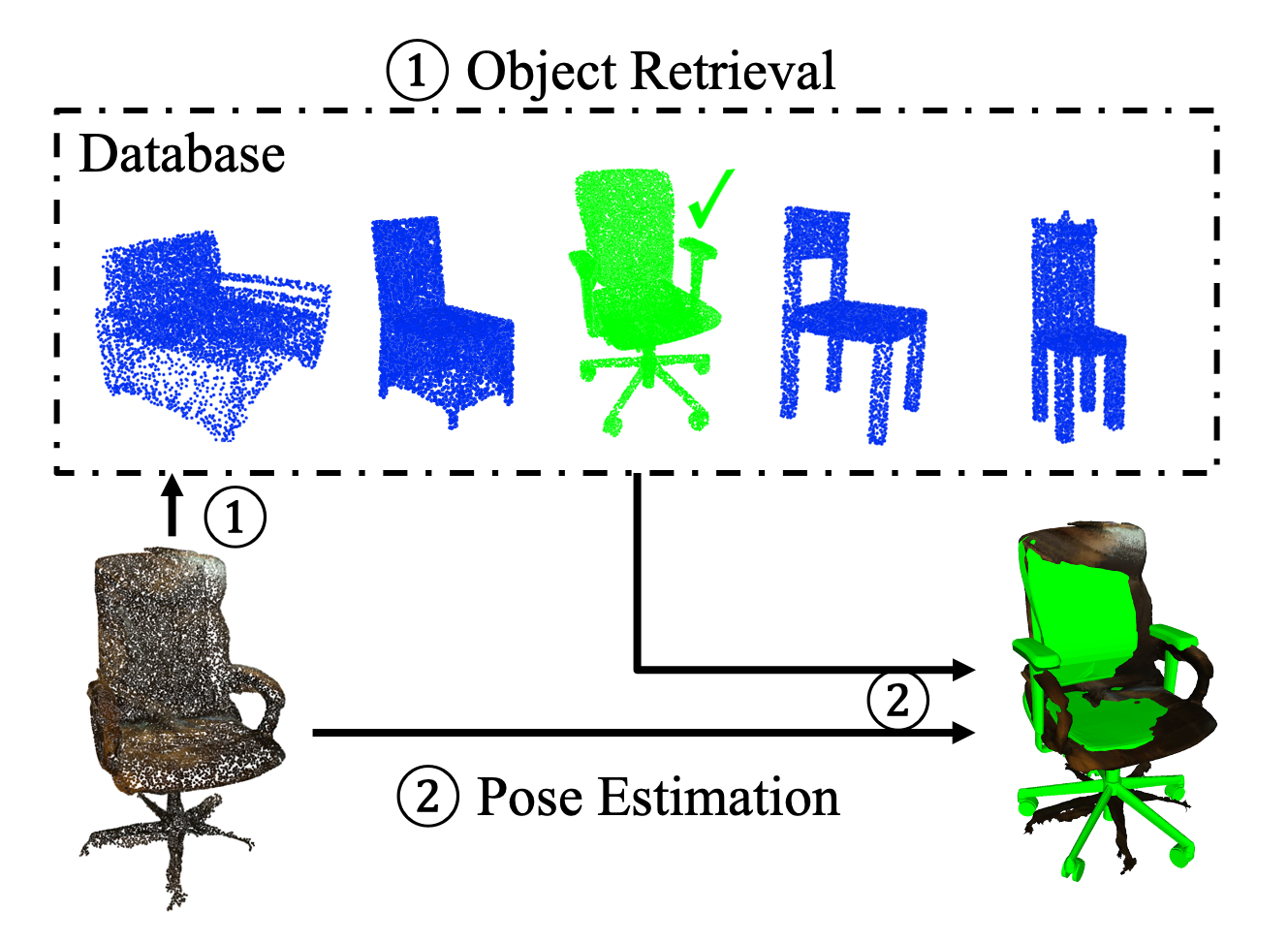}
  \caption{Given an observed object point-cloud, we focus on retrieving a similar object (green) from a category database (blue) and estimating its pose with respect to the input object. Performing retrieval and registration for object point-clouds observed online allow us to construct an object-level map of an unknown environment.}
  \label{fig:overview}
\end{figure}

Most existing work focuses on category-level object retrieval or object retrieval with fixed poses \cite{Esteves_2018_learning,He_2018_triplet,Uy_Deformation_2020} or pose estimation between identical object pairs \cite{Choy_FCGF_ICCV2019,Choy_DGR_2020,Yang_TEASER_2020}, differing only due to noise or occlusion but not due to shape variation. This paper considers object pose estimation from partial point-cloud observations to enable online object-level mapping. We assume that an object detection and segmentation model, trained offline using a large database of images or object CAD models, is available to segment object instances across multiple camera views and provide partial point-cloud observations. For each observed instance, we focus on retrieving a similar CAD model from the offline shape database and aligning it to the observation to estimate the observed object's pose. Fig.~\ref{fig:overview} illustrates the problem setting.

Given a query point-cloud, we extract global features to enable retrieval and local point-wise features to enable point-cloud matching and pose registration. We present CORSAIR, an approach for fully convolutional object retrieval and symmetry-aided registration. CORSAIR extends the Fully Convolutional Geometric Features (FCGF) model \cite{Choy_FCGF_ICCV2019} by introducing a latent code embedding architecture, which provides a global object-level feature in addition to the local point-wise features. The local point-wise features are trained using cross-object point matches in normalized canonical coordinates \cite{Wang_NOCS_CVPR2019}. The global feature is extracted from a hierarchical abstraction of the bottleneck layer of the FCGF encoder-decoder architecture. Both are trained using metric learning with contrastive loss for the local features and triplet loss for the global feature. Our formulation also leverages symmetries in the object shapes to obtain promising feature matches in each symmetric class, greatly improving pose estimation. Our \textbf{contributions} are summarized as follows.
\begin{itemize}
    \item We design a sparse fully convolutional network to jointly regress global and local point-cloud features, which are hierarchically correlated. The global feature enables similar model retrieval, while the local features allow object pose registration.
    
    \item We construct symmetry classes within an object instance based on the local features and aid the generation of promising feature pairs for robust registration.
    
    
    
\end{itemize}
The effectiveness of these contributions is validated in both synthetic and real-world object datasets, containing different object categories.

\section{Related Work}
\label{sec:related_work}

Deep neural network techniques for object detection and semantic segmentation \cite{He_Mask_2020,Gao_Res2Net_2021} have reached impressive levels of performance. They have allowed traditional dense geometric simultaneous localization and mapping (SLAM) approaches \cite{Whelan_ElasticFusion_2016,Dai_BundleFusion_2017} to integrate semantic content in the scene reconstruction \cite{Tateno_CNN-SLAM_2017,McCormac_SemanticFusion_2017}. Since many robotics applications involve object interaction, SLAM algorithms which provide sparse object-level reconstruction \cite{Salas-Moreno_slam++_2013,Bowman_SemSLAM_2017,Runz_MaskFusion_2018,Mccormac_Fusion++_2018} instead of dense maps play an important role as well.

While object detection, segmentation and tracking are already well studied, we focus on object pose estimation from partial point-clouds obtained online for object-level mapping. To estimate poses of unknown instances, we rely on registration of known instances from the same category with respect to the observed point-cloud. The availability of massive object CAD datasets \cite{Chang_shapenet_2015} makes it possible to select a similar-looking instance from the database to increase the accuracy of estimating the pose of the unknown instance. Hence, object retrieval is an important sub-problem for robust pose registration.
Compared to category classification of 3D objects \cite{Esteves_2018_learning,He_2018_triplet}, retrieval of specific CAD instances is more challenging due to the emphasis on shape similarity. Grabner et al. \cite{Grabner_3DPose_2018} render CAD model depth and embed the depth and RGB image observations jointly for CAD model retrieval. Dahnert et al. \cite{Dahnert_Joint_2019} use a 3D hourglass encoder-decoders structure to learn an embedding feature with triplet loss for shapes, implicitly represented using a signed distance field. Uy et al. \cite{Uy_Deformation_2020} introduce a deformation-aware asymmetric distance across CAD models and learn an egocentric anisotropic distance field for latent embeddings. Most of these works, however, perform retrieval with canonical object poses and, hence, do not consider pose invariance or registration. On the other hand, many point-cloud registration approaches assume the point-clouds are from the same object or scene. DGR \cite{Choy_DGR_2020} predicts correspondence confidence of 3D point pairs using a 6D convolutional network and applies a weighted Procrustes algorithm, making the whole process differentiable. TEASER \cite{Yang_TEASER_2020} is a certifiable registration algorithm handling high outlier rates with a truncated least squares formulation and semidefinite relaxation. 

The Scan2CAD dataset \cite{Avetisyan_Scan2CAD_2019} annotates 6D pose and scale of objects in the indoor scenes of ScanNet \cite{Dai_ScanNet_CVPR2017} by aligning CAD models from ShapeNet. RGBD scans are converted to voxelized signed distance fields and a 3D CNN network is used to predict sparse keypoint correspondence, given a matching CAD model. NOCS \cite{Wang_NOCS_CVPR2019} uses the idea of normalized canonical coordinates for a specific category and generates dense annotations covering the whole object surface. This model can predict the normalized canonical coordinates densely on a query image and use them to recover the object pose. Feng et al. \cite{Feng_CatFCGF_2020} align different object instances from the same category in normalized canonical coordinates and learn cross-instance matching FCGF features. Mask2CAD \cite{Kuo_Mask2CAD_2020} detects and segments objects in a single RGB image, after which a CAD model is retrieved and its pose is regressed. The CAD model embedding is generated by rendering 2D images from different views to overcome the modality gap. Vid2CAD \cite{Maninis_Vid2CAD_2020} leverages multi-view consistency constraints to resolve scale and depth ambiguities so as to derive a temporally consistent pose estimation of the objects.

We extend the work of \cite{Feng_CatFCGF_2020}, which tackles intra-category point-cloud matching by enabling retrieval of a similar instance from the category database for matching. We show that a global object-level feature for retrieval can be generated hierarchically from the local point-wise FCGF features for matching and alignment. We also leverage the symmetry of artificial objects to derive a more robust pose estimation approach.

\section{Problem Formulation}
\label{sec:problem_formulation}

Consider a robot, equipped with an RGBD camera, aiming to construct an object-level map of an unknown environment. Assume that the camera pose is estimated using an odometry algorithm, such as ORB-SLAM3 \cite{Campos_ORB3_2020}. Assume also that a convolutional neural network, such as Mask R-CNN \cite{He_Mask_2020}, is used to detect and segment objects in each RGB image, and an object tracking algorithm, such as FairMOT \cite{Zhang_FairMOT_2020}, tracks the object detections over time. A partial point-cloud observation $\bfX \in \bbR^{3 \times N}$ of a tracked object instance can be obtained by accumulating the segmented RGBD pixels associated with the instance over time and projecting them to the world frame using the estimated camera pose trajectory.

Let $\calY := \crl{\bfY_i \in \bbR^{3 \times M_i}}_i$ be a database of point-cloud object models from the same category as $\bfX$. We assume the database was used offline for training the object detection and tracking models and is available to the robot. We consider the following joint object retrieval and registration problem.

\begin{problem}
\label{problem1}
Given a query point-cloud $\bfX \in \bbR^{3 \times N}$ and a point-cloud database $\calY := \crl{\bfY_i \in \bbR^{3 \times M_i}}_i$, retrieve a point-cloud $\bfY \in \calY$ that is similar to $\bfX$ and estimate its rotation $\bfR \in SO(3)$ and translation $\bfp \in \bbR^3$ with respect to $\bfX$:
\begin{equation}
\label{eq:problem}
    \min_{\bfY\in\calY,\bfR \in SO(3), \bfp \in \bbR^3} d(\bfX, \bfR \bfY + \bfp \mathbf{1}^\top),
\end{equation}
where $\mathbf{1}$ is a vector with all elements equal to $1$ and $d$ is a point-cloud distance metric. 
\end{problem}

There are different ways to specify a point-cloud distance $d$ in Problem~\ref{problem1}. We measure the average distance between matching points $\bfx_i \in \bbR^3$ in $\bfX$ and $\bfy_{m(i)} \in \bbR^3$ in $\bfY$:
\begin{equation}
\label{eq:d}
    d(\bfX,\bfY) = \frac{1}{N}\sum_{i=1}^N \|\bfy_{m(i)} - \bfx_i\|^2\indicator_{\crl{m(i) \neq 0}},
\end{equation}
where $m:\crl{1,\ldots N} \mapsto \crl{0,1,\ldots,M}$ associates the indices $i$ of the points in $\bfX$ with the indices $j = m(i)$ of the points in $\bfY$, and $m(i)=0$ indicates that $i$ does not match any index in $\bfY$.

The objective of Problem~\ref{problem1} is to determine the world-frame pose of an object instance, observed online, which may or may not have been seen before. Retrieving a similar instance from the training database and registering it with the point-cloud observation allows accurate pose estimation of the newly observed object.

\section{Approach}
\label{sec:approach}

\begin{figure}[t]
  \includegraphics[width=\linewidth]{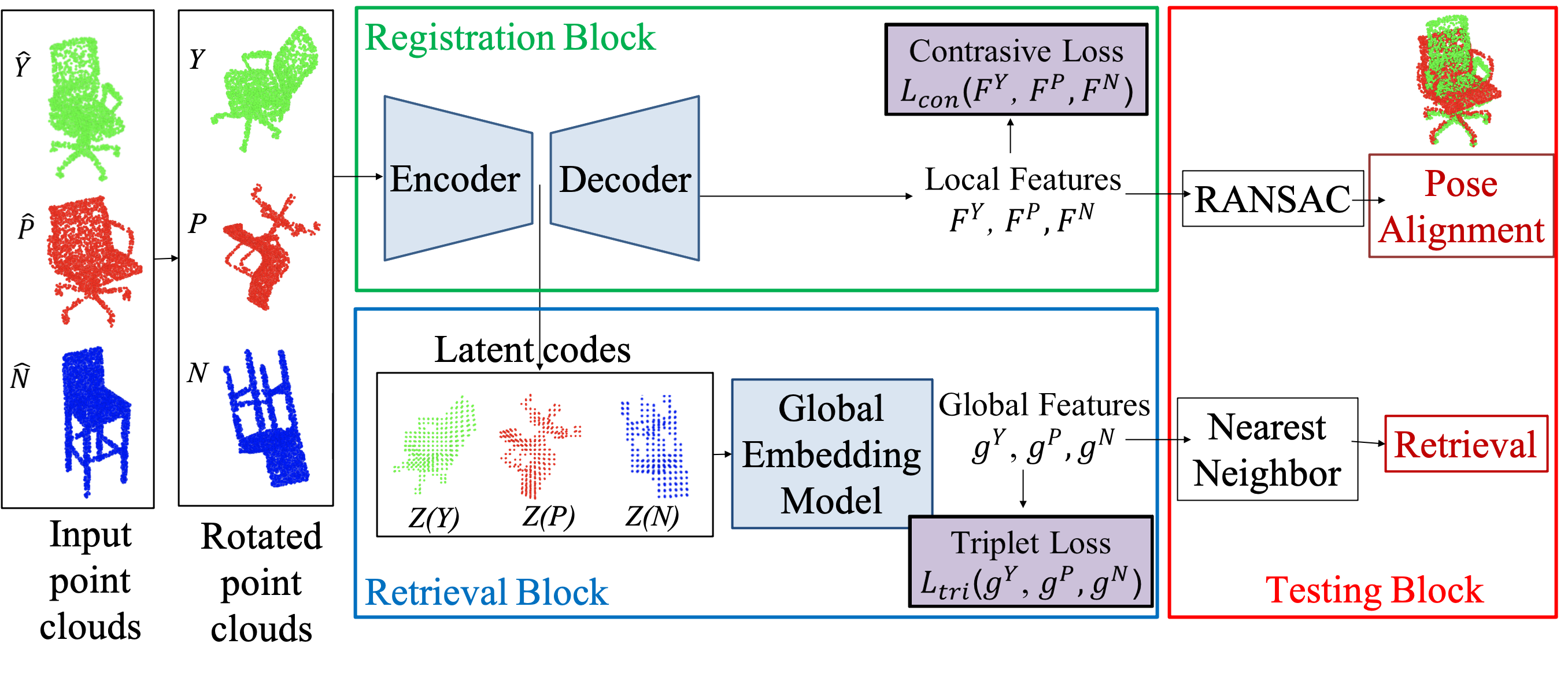}
  \caption{During training, given a point-cloud and its corresponding positive and negative pairs $(\bfY, \bfP,\bfN)$, a Registration block is trained to generate local point-wise features (Sec. \ref{sec:registration}) and a Retrieval block is trained to generate a global shape embedding (Sec. \ref{sec:retrieval}). During testing (Sec. \ref{sec:inference}), given a query point-cloud $\bfX$, we generate its global embedding $\bfg^\bfX$ and retrieve a similar instance using nearest neighbors in the embedding space. Local features are then generated for both point-clouds, and matching pairs are used to recover the pose of the query using RANSAC.}
  \label{fig:pipeline}
\end{figure}

Estimating the pose of novel objects based on a finite set of models from the same category can be challenging due to shape variation. We extend the point-wise FCGF feature extractor proposed in \cite{Choy_FCGF_ICCV2019} with a global embedding network. We learn \emph{local point-wise features} (Sec. \ref{sec:registration}) to enable robust matching and registration of point-cloud with potentially different shapes. We learn \emph{global object-level features} (Sec. \ref{sec:retrieval}) to enable retrieval of a point cloud from the database that is similar to the query point cloud. During inference (Sec. \ref{sec:inference}), we align a partially observed point-cloud with a retrieved one, and exploit object symmetry to generate matching feature pairs for registration. Fig.~\ref{fig:pipeline} presents an overview of our approach. 



\subsection{Local Features for Pose Registration} 
\label{sec:registration}




We aim to predict matching pairs of point-wise local feature for pose registration between point clouds. During training, we first define matching pairs of points, rather than features. If two point-clouds $\mathbf{X}\in \bbR^{3 \times N}$ and $\mathbf{Y}\in \bbR^{3 \times M}$ were already aligned in the same coordinate frame, matching point pairs can be extracted via:
\begin{equation} \label{eq:pairs}
    \scaleMathLine[0.9]{p(\mathbf{X},\mathbf{Y}) = \{(i,j) \in \bbN^2 \mid \|\mathbf{x}_i-\mathbf{y}_j\|<\tau, i \leq N, j \leq M\},}
\end{equation}
where $\tau > 0$ is a matching tolerance. Negative pairs can be obtained from the complement set of the positive pairs, ensuring that two negatively associated points are at least a margin $\tau$ away.

Since the training set $\calY$ contains point clouds from different instances, which inherently carry geometric shape differences, we generate matching pairs in category-level normalized canonical coordinates (NCCs) \cite{Wang_NOCS_CVPR2019,Feng_CatFCGF_2020}. However, instead of dense correspondence annotation as done in \cite{Wang_NOCS_CVPR2019,Feng_CatFCGF_2020}, we only need object pose annotations to convert the point-clouds to NCCs. Given the scale $s_{\bf{X}}\in\bbR$, rotation $\bfR_\bfX \in SO(3)$, and translation $\bfp_\bfX \in \bbR^3$ of a point-cloud $\bfX$ during training, $\bfX$ can be converted to NCCs via:
\begin{equation}
    \mathbf{X}_{\text{NCC}} = s^{-1}_{\bf{X}} \cdot \bfR^\top_\bfX \prl{\bfX - \bfp_\bfX \mathbf{1}^\top}.
    \label{eq:NCC}
\end{equation}
Thus, to generate matching pairs for different instances $\bfX$ and $\bfY$ of the same category, we convert both into NCCs, and obtain the positive pair set as $\mathcal{P_L} = p(\bfX_\text{NCC},\bfY_\text{NCC})$. A negative pair set $\mathcal{N_L}$ is obtained as a subset of the complement of $\mathcal{P_L}$. 

\begin{figure}[t]
  \includegraphics[width=\linewidth]{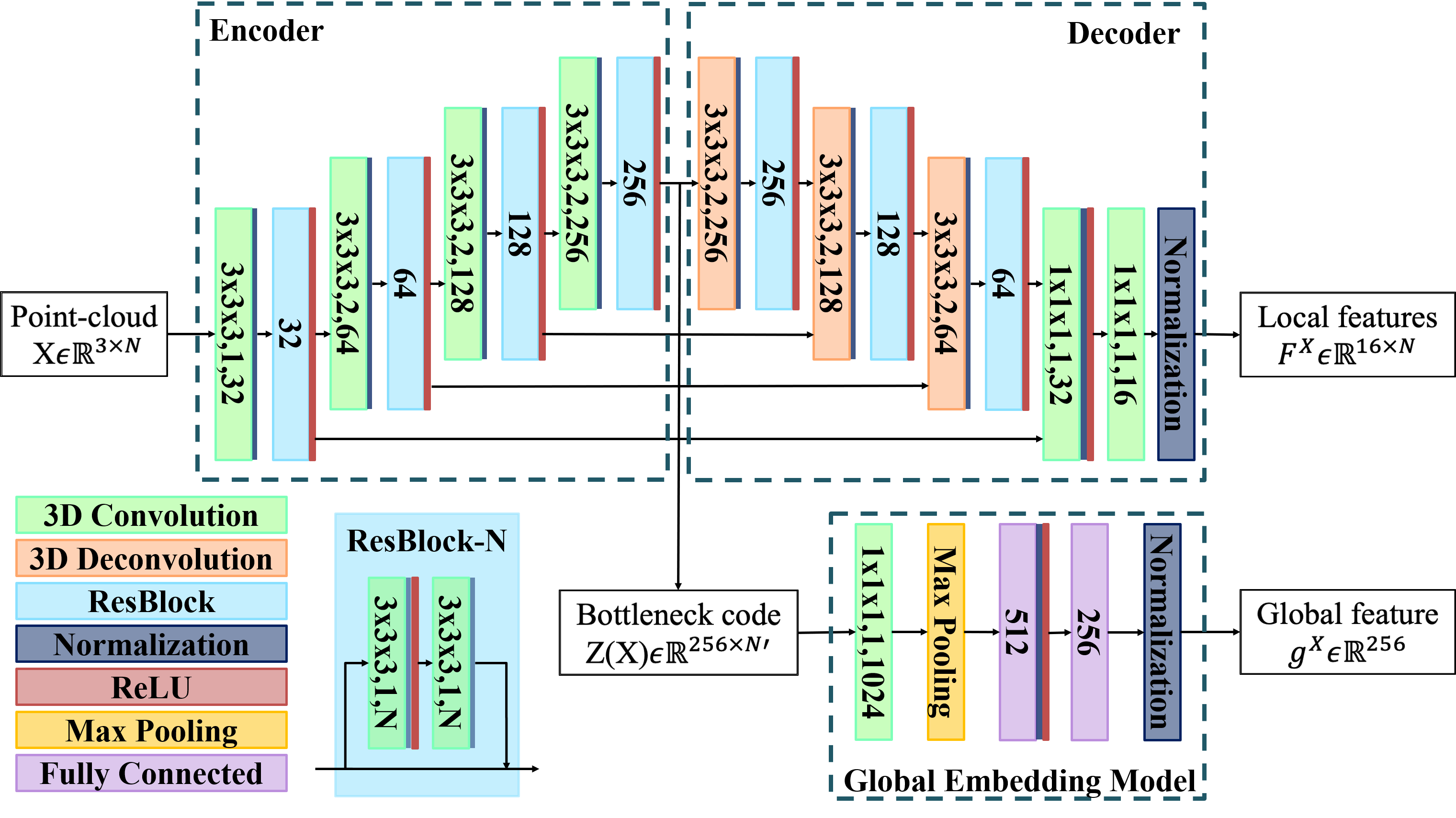}
  \caption{Our model extends the sparse convolutional encoder-decoder ResUNet structure proposed in FCGF \cite{Choy_FCGF_ICCV2019} by adding an embedding module to the bottleneck latent code (encoder output). The output of our embedding module provides a global object shape feature, suitable for retrieval, while the decoder generates point-wise local features. The numbers in the 3D convolution and deconvolution blocks represent kernel sizes, strides, and output dimensions. The numbers in the fully connected blocks represent the output dimensions. Each ResBlock is composed of two 3D convolution layers and the number indicates the output dimensions.}
  \label{fig:network}
\end{figure}

Given a point-cloud $\bfX \in \bbR^{3 \times N}$, our model uses a sparse fully convolutional encoder-decoder architecture, illustrated in Fig. \ref{fig:network}, to predict local point-wise features:
\begin{equation}
    \bfF^\bfX = \left[\bff^\bfx_1,\ldots,\bff^\bfx_N\right] \in \bbR^{C \times N},
\end{equation}
where $\bff^\bfx_i$ is the feature
corresponding to the point $\bfx_i$.
Sparse convolution \cite{Choy_4D_2019} generalizes image convolution to arbitrary dimensions and coordinates and allows processing of spatially sparse inputs. Our model is an extension of the FCGF model \cite{Choy_FCGF_ICCV2019} that adds an embedding module to the encoder output (bottleneck layer) to also retrieve a global feature. The training and role of the global feature for retrieval is described in Sec.~\ref{sec:retrieval}. 

We use metric learning to train the local feature extractor. Relying on the positive pairs $\mathcal{P_L}$ and the negative pairs $\mathcal{N_L}$ of matching points, we define a contrastive loss function for the features $\bfF^\bfX$ and $\bfF^\bfY$ associated with two point clouds from the training set: 
\begin{equation}
  \begin{aligned}
L_{\text{con}}(\bfF^\bfX,\bfF^\bfY)=& \sum_{(i,j)\in\mathcal{P_L}} \max(0,\|\bff^\bfx_i-\bff^\bfy_j\|_2-p_+)^2\\
+ &\sum_{(i,j)\in\mathcal{N_L}} \max(0,p_{-}-\|\bff^\bfx_i-\bff^\bfy_j\|_2)^2,
  \end{aligned}
  \label{eq:contrasive_loss}
\end{equation}
where $p_+$ and $p_-$ are the positive and negative thresholds. These thresholds are selected to ensure that points from the positive pairs move closer together and points from the negative pairs move farther apart in the feature space. We normalize the feature vectors to unit length and set $p_+ = 0.1$ and $p_-=1.5$.

For each point cloud $\bfY$ in the training set $\calY$, we choose similar point clouds $\bfP$ and dissimilar point clouds $\bfN$ defined precisely in Sec.~\ref{sec:retrieval}. We, then, generate the positive pairs between $\bfY$ and $\bfP$ using \eqref{eq:pairs} as $\mathcal{P_L} = p(\bfY_{NCC},\bfP_{NCC})$. We sample the negative pair set $\cal{N_L}$ by taking random pairs between $\bfY$ and $\bfN$ as well as from the complement of $\mathcal{P_L}$. The local feature extractor model is trained with the contrastive loss in \eqref{eq:contrasive_loss} but using the pairs from $\cal{P_L}$ and $\cal{N_L}$.

%
%

%
%

%




\subsection{Global Feature for Object Retrieval}
\label{sec:retrieval}

Extracting a similar point cloud from $\calY$ for a given query point cloud $\bfX$ is crucial for pose registration because only similar shapes provide consistent local geometric features for matching. Since the sparse convolutional model in Fig.~\ref{fig:network} performs in providing local geometric features, we design a global shape descriptor by combining the local features. An input point cloud $\bfX$ is quantized into a sparse tensor and gets downsampled by the encoder into a point cloud $\bfZ(\bfX) \in \bbR^{256 \times N'}$ with fewer points $N' < N$ but each in a higher ($256$ in our case) dimension. After the multiple convolution layers of the encoder, the bottleneck code $\bfZ(\bfX)$ encodes a high-level structure feature which should be beneficial for shape retrieval. We introduce an embedding module to extract a single global feature $\bfg^\bfX \in \bbR^{256}$ from $\bfZ(\bfX)$ as $\bfg^\bfX = g(\bfZ(\bfX))$. As shown in Fig.~\ref{fig:network}, the embedding module $g(.)$, includes a fully convolutional layer, followed by maxpooling to combine the features from all points in $\bfZ(\bfX)$ and pass them through several fully connected layers to obtain $\bfg^\bfX$.




We use metric learning for global feature training too. To measure the similarity of point cloud $\bfX \in \bbR^{3 \times N}$ with respect to $\bfY \in \bbR^{3 \times M}$, we define a Single-direction Chamfer Distance (SCD):
\begin{equation}
d_{SCD}\left(\mathbf{X,Y}\right)=\sum_{i = 1}^N \min_{j = 1}^M\|\bfx_i-\bfy_j\|_{2}^{2}.
\label{eq:chamfer_single}
\end{equation}
The bi-directional similarity between the two point clouds is measured by the usual Chamfer distance:
\begin{equation}
d_{CD}\left(\mathbf{X,Y}\right)=d_{SCD}\left(\mathbf{X,Y}\right) + d_{SCD}\left(\mathbf{Y,X}\right)
\label{eq:chamfer}
\end{equation}
Let $\mathbf{D} \in \mathbb{R}^{|\calY| \times |\calY|}$ encode the pair-wise Chamfer distance similarity of all point clouds in $\calY$ with elements:
\begin{equation}
    \mathbf{D}_{i,j} = d_{CD}(\bfY_i, \bfY_j), \;\;\text{for } \bfY_i, \bfY_j \in \mathcal{Y}.
    \label{eq:D_matrix}
\end{equation}
A similarity ranking for $\bfY_i \in \calY$ can be obtained by sorting the $i$-th column of $\mathbf{D}$ in ascending order. We define a positive set $\mathcal{P_G}(\bfY_i)$ and negative set $\mathcal{N_G}(\bfY_i)$ of point clouds associated with $\bfY_i$ as follows:
\begin{align*}
\mathcal{P_G}(\bfY_i)&=\{\mathbf{Y}_j \mid \text{Rank}_i(\bfY_j) \leq \tau_+|\calY|, d_{CD}(\bfY_i, \bfY_j) \leq \delta_+ \}\\
\mathcal{N_G}(\bfY_i)&=\{\mathbf{Y}_j \mid \text{Rank}_i(\bfY_j) \geq \tau_-|\calY|, d_{CD}(\bfY_i, \bfY_j) \geq \delta_- \}
\end{align*}
%
where $\text{Rank}_i(\bfY)$ returns an integer indicating the Chamfer distance ranking of $\bfY$ to $\bfY_i$, $\tau_+$ and $\tau_-$ are the percentage of positive and negative point clouds we want to consider, and $\delta_+$ and $\delta_-$ are the Chamfer distance thresholds. In our experiments, we set $\tau_+=0.1$, $\tau_-=0.5$, $\delta_+=0.15$ and $\delta_-=0.20$.

For a given point-cloud $\bfY$, we randomly select one positive object $\mathbf{P} \in \mathcal{P_G}(\bfY)$ and one negative object $\mathbf{N} \in \mathcal{N_G}(\bfY)$ and train the global embedding module with a triplet loss:
\begin{equation}
\label{eq: global triplet loss}
    L_{\text{tri}}(\bfg^\bfY,\bfg^\bfP,\bfg^\bfN) = \max(1 + \|\bfg^\bfY-\bfg^\bfP\|_2 - \|\bfg^\bfY- \bfg^\bfN\|_2, 0).
\end{equation}
Similar to the contrastive loss for the local features, the triplet loss pushes similar point-clouds closer and drags dissimilar point-clouds apart in the global embedding space. 



\subsection{Inference}
\label{sec:inference}

\begin{figure}[t]
  \includegraphics[width=\textwidth/2,trim=0 5 0 5,clip]{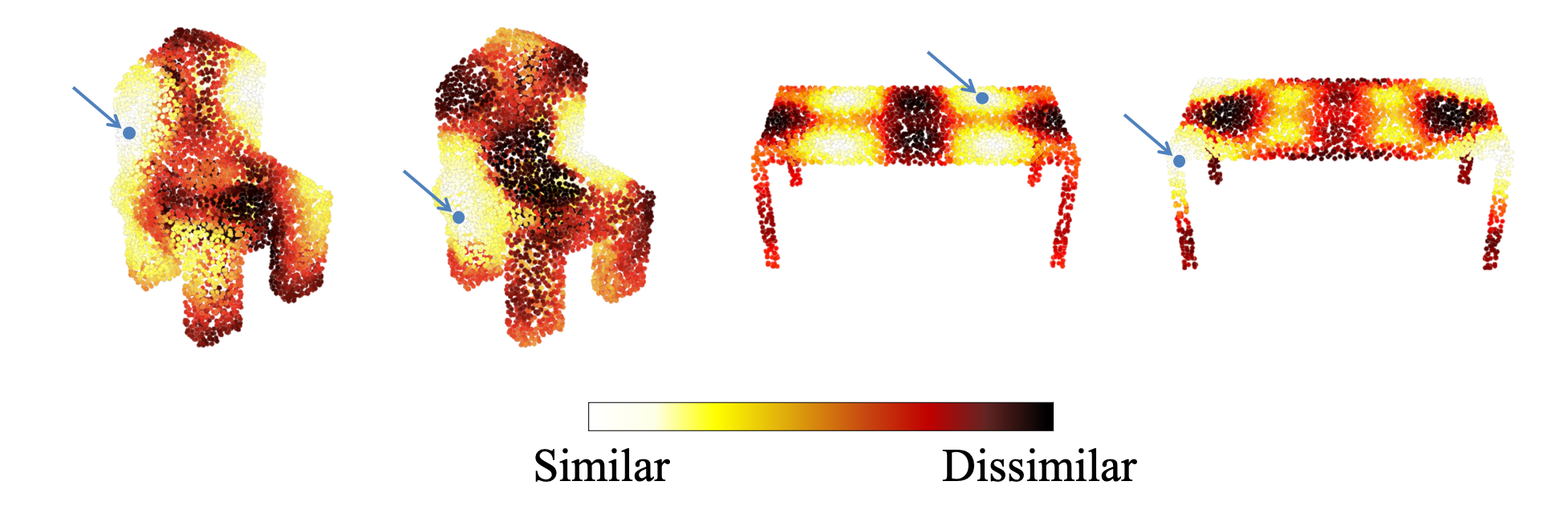}
  \caption{Heatmap of local feature similarity for different points on a chair (left) and a table (right) instances. Lighter points indicate points with similar features to the query point (blue dot), while darker ones have dissimilar features. The feature similarity heatmap visualizes the symmetric nature of the local features.}
  \label{fig:heatmap}
\end{figure}

Finally, we consider pose registration for a query point cloud $\bfX$ given the trained CORSAIR model in Fig.~\ref{fig:network}. The first step is to retrieve a similar model from $\calY$. The local and global features of all point clouds in $\calY$ are pre-computed offline. The query $\bfX$ is passed through the CORSAIR model to obtain its local features $\bfF^\bfX$ and global feature $\bfg^\bfX$. The instance from $\bfY$ closest to $\bfX$ is retrieved via:
\begin{equation}
\label{eq:retrieval}
    \bfY = \argmin_{\bf{Y_i}\in\calY} \|\bfg^\bfX - \bfg^{\bf{Y_i}}\|_2,
\end{equation}
Next, we generate matching pairs between $\bfX$ and $\bfY$ by searching for $K$ nearest neighbors in the local feature space:
\begin{align}
    nn_K(\mathbf{F}^{\mathbf{X}},\mathbf{F}^{\mathbf{Y}}) = \{(i,j_i) &| j_i\in \text{$K$-$\argmin_j$} \|\mathbf{f}^\mathbf{x}_i-\mathbf{f}^\mathbf{y}_j\|_2, \notag\\
    &i \leq N, j_i \leq M, i,j_i \in \mathbb{N}\} \label{eq:feature_pairs} 
\end{align}
where we define $K$-$\arg\min_j$ as the set of $K$ different values that make the function smaller than any other set of $K$ indices. The correspondence candidates in $nn_K(\mathbf{F}^{\mathbf{X}},\mathbf{F}^{\mathbf{Y}})$ are used to recover the rotation and translation of $\bfY$ with respect to $\bfX$ via a robust pose estimation method such as RANSAC \cite{Fischler_RANSAC_1987}.


Artificial objects usually have one or more planes of symmetry. Since our model is able to generate rotation-invariant local features, the features from symmetrical areas can be similar, which significantly increases the risk of mismatch in \eqref{eq:feature_pairs}. A feature-distance heatmap shown in Fig.~\ref{fig:heatmap} illustrates the symmetrical patterns. We propose a symmetry-aware method to add constraints in the nearest neighbor matching phase.
We split the point-cloud with a symmetric segmentation method (see Alg.~\ref{approach: symmetry algorithm}) and then constrain the nearest-neighbor matching to the corresponding parts. 

\begin{figure}[t]
\centering
  \includegraphics[height = 3cm]{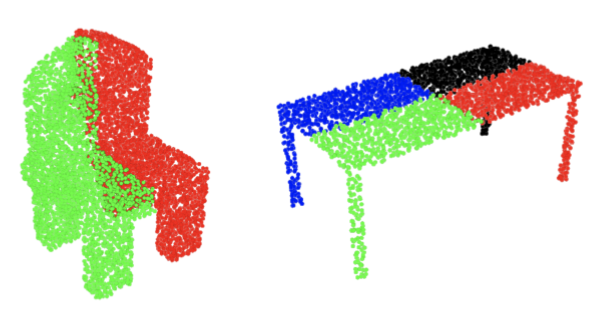}
  \caption{Different objects split according to their symmetry planes.}
  \label{fig:symmetry-visualization}
\end{figure}

Given an object category, we assume that the number of symmetric classes $G$, computed as the number of symmetry planes times $2$, is known. For example, $G=2$ for chairs and $G=4$ for tables. We first extract the point-wise local features $\mathbf{F}^\bfX$ for the input point-cloud $\bfX$. Second, we randomly sample ${n}$ points from the point-cloud. For each sampled point, we take its $K$ nearest neighbors, $nn_K(\mathbf{f}^\bfX_i, \mathbf{F}^\bfX)$, and perform $G$-means clustering using their 3D spatial coordinates. The object can then be split into $G$ parts, $\{\mathbf{\widetilde{X}}_1,\dots,\mathbf{\widetilde{X}}_G\}$, by the decision boundaries of $G$-means clustering as shown in Fig. \ref{fig:symmetry-visualization}. Each of the splits is considered as a candidate and we choose the most even split as our symmetric segmentation output. The evenness of a split is measured by the standard deviation $\sigma$ of the sizes of the $G$ parts. 

We assume that the query and retrieved point clouds $\bfX$ and $\bfY$ share the same symmetry property and split them with our symmetry segmentation method. Since there are multiple possible mappings between the subsets $\{\mathbf{\widetilde{X}_1},\dots,\mathbf{\widetilde{X}}_G\}$ and $\{\mathbf{\widetilde{Y}_1},\dots,\mathbf{\widetilde{Y}}_G\}$, we generate matching pairs using \eqref{eq:feature_pairs} for all the possible mappings. We also generate matching pairs without the symmetry constraints as a back-up for asymmetric objects. We supply these sets of matching pairs to RANSAC to estimate the rotation $\mathbf{R}$ and translation $\mathbf{p}$ that align $\bfY$ with $\bfX$ for every possible matching pair set. The quality of alignment is evaluated by the single direction Chamfer distance $d_{SCD}(\bfX, \bfR \bfY + \bfp \mathbf{1}^\top)$ defined in \eqref{eq:chamfer_single}. The rotation and translation with the best alignment quality are selected as the output of our symmetry-aware pose estimation method. Our symmetry-aware method performs well when the input point-clouds have symmetrical structure.

\begin{algorithm}[t]
\begin{algorithmic}[1]
\small
\State \textbf{input}: point-cloud $\mathbf{X}$, point-wise local features $\mathbf{F}^\bfX$, number of matching feature pairs $M$, number of symmetry classes $G$
\State Randomly sample $\mathcal{S} \leftarrow \{\bfx_i| \bfx_i \in \mathbf{X}, i \leq n \}$ 
 \For{$ \bfx_i \in \mathcal{S}$}
     \State $\mathcal{K} \leftarrow \{ \bfx_j | (i, j) \ \in nn_K(\bff^\bfx_i, \mathbf{F}^\bfX) \}$
     \State Split $\mathbf{X}$ in clusters $\calC_i =\{\widetilde{\bfX}_1,\ldots,\widetilde{\bfX}_G\}$ via $\Call{GMeans}{\mathcal{K}}$
     \State $\sigma_i \gets$ Standard Deviation of $\{|\widetilde{\bfX}_1|,\ldots,|\widetilde{\bfX}_G|\}$ 
 \EndFor
 \State \textbf{output}: $\calC_i$ with the smallest $\sigma_i$
\caption{Symmetry-aided Segmentation}
\label{approach: symmetry algorithm}
\end{algorithmic}
\end{algorithm}

\begin{figure}[t]
  \includegraphics[width=\textwidth/2,trim=0 50 0 50,clip]{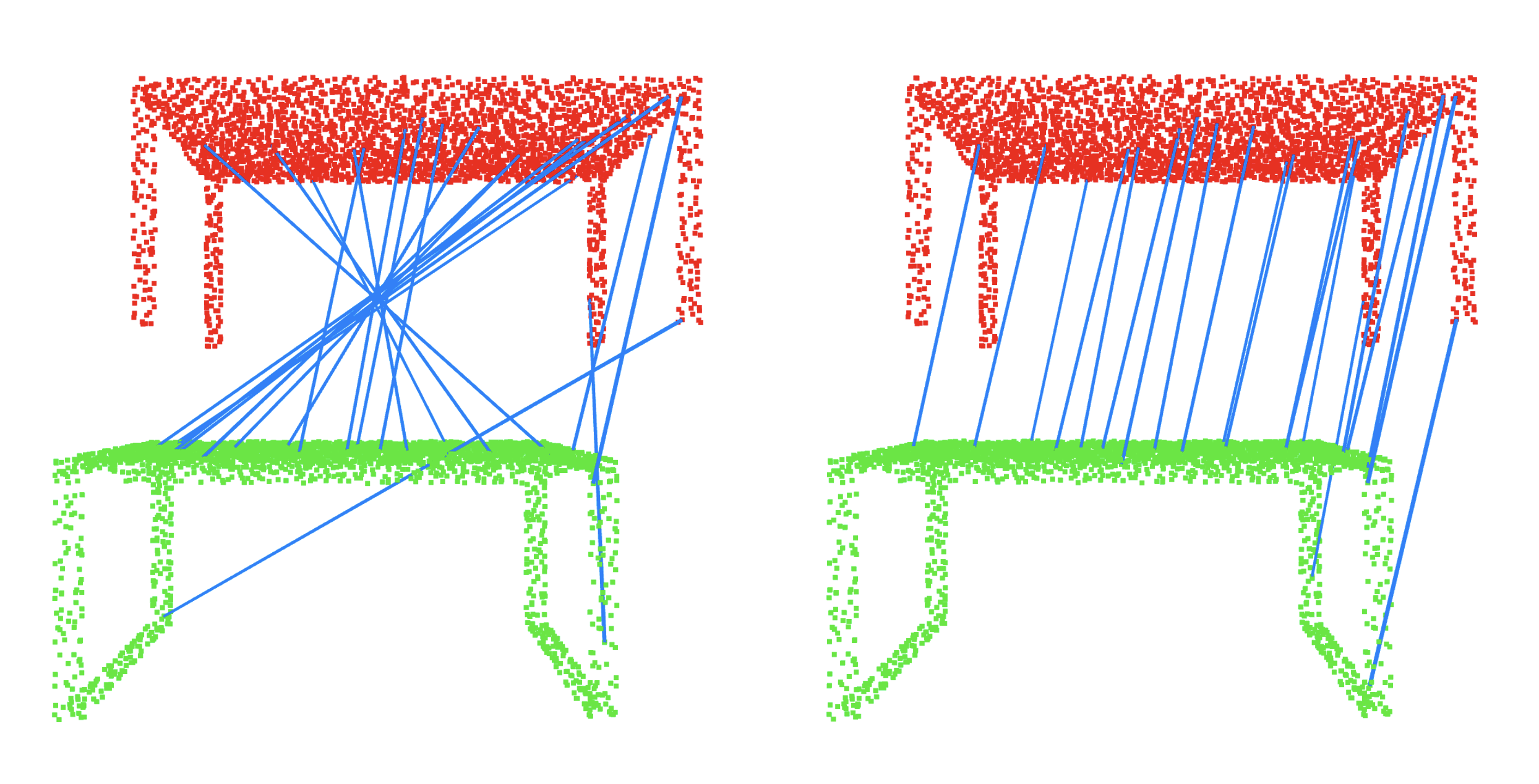}
  \caption{Local feature matching between a query point-cloud (red) and a retrieved point-cloud (green). The blue lines show the pairs between the two point-clouds based on nearest neighbor matching of the local features. The left pair of point-clouds shows matching without symmetry aid, while the right pair shows matching after detecting the plane of symmetry.}
  \label{fig:matching}
\end{figure}

\section{Evaluation}
\label{sec:experiments}


We evaluate the performance of CORSAIR on the synthetic ShapeNet \cite{Chang_shapenet_2015} dataset and the real-world Scan2CAD \cite{Avetisyan_Scan2CAD_2019} dataset. We assume that the category of any given point-cloud is known so that we can train models for different categories separately. In our experiments, chair and table are selected. 

\subsection{Evaluation Metrics}
\label{sec:metric}

Given a query point-cloud $\bfX$ and its retrieved point-cloud $\bfY$, we define the ground truth pose of $\bfX$ with respect to $\bfY$ as $\mathbf{(R^*,p^*)}$ and the estimated pose as $\mathbf{(\hat{R},\hat{p})}$.
We compute relative rotation error (RRE) and relative translation error (RTE) as pose estimation metrics:
\begin{equation}
\mathbf{RTE(\mathbf{\hat{p}},\mathbf{p^*})} = ||\mathbf{\hat{p}} - \mathbf{p^*}||_{2},
\end{equation}
\begin{equation}
\mathbf{RRE(\mathbf{\hat{R}},\mathbf{R^*})} = \arccos((\mathbf{tr(\hat{R}^{\top}R^*}) -1)/ 2).
\end{equation}
Two different metrics are used to evaluate the retrieval performance on ShapeNet: Precision@M and Top-1 Chamfer distance. Precision@M is defined as:
\begin{equation}
    \text{Precision@M} = \frac{1}{M}\sum_{\mathbf{Y} \in \mathcal{R}} \indicator_{\mathbf{Y} \in \mathcal{P_G}(\bfX)}, 
\end{equation}
where $\mathcal{R} \subset \mathcal{Y}$ is the retrieved set of M CAD models and $\mathcal{P_G}(\bfX)$ is the ground truth positive set for query object $\mathbf{X}$ as defined in \ref{sec:retrieval}. 
The Top-1 Chamfer distance metric is calculated by measuring the Chamfer distance \eqref{eq:chamfer} between the top-1 retrieved CAD model and the ground-truth top-1 similar CAD model.

\subsection{ShapeNet}

\begin{table*}[]
\centering
\caption{Quantitative pose registration results in ShapeNet \cite{Chang_shapenet_2015}. Every object is aligned with a randomly selected object from its top-10\% similar positive set. RANSAC is applied for all the methods for pose registration. The table shows the percentage of test cases below different error thresholds for different feature extraction and matching strategies.}
\begin{tabular}{c|c|ccc|ccc}
\hline
Category & Method & RRE$\leq 5$º & RRE$\leq 15$º & RRE$\leq 45$º & RTE$\leq 0.03$ & RTE$\leq 0.05$ & RTE$\leq 0.10$ \\ 
\hline
\multirow{3}{*}{Chair}  & FPFH \cite{Rusu_FPFH_2009} &      
1.4 & 7.8 & 18.5 & 2.0 & 10.7 & 25.8 \\ 
  & FCGF \cite{Choy_FCGF_ICCV2019} &      
41.0 & 86.1 & 96.1 & 32.2 & 83.8 & 96.0 \\ 
                       &  CORSAIR (Ours)  & 
62.3 & 92.1 & 98.1 & 48.5 & 90.2 & 97.7 \\ 
\cline{1-8}
\multirow{3}{*}{Table}  & FPFH \cite{Rusu_FPFH_2009}  &    
3.8 & 18.2 & 34.4 & 3.0 & 14.9 & 32.5 \\ 
 & FCGF \cite{Choy_FCGF_ICCV2019} &    
25.4 & 62.7 & 78.4 & 24.9 & 63.8 & 80.5 \\ 
                       &   CORSAIR (Ours)   &
54.5 & 74.8 & 82.7 & 46.0 & 77.0 & 85.0 \\ 
\hline
\end{tabular}
\label{table:pose_shapenet}
\end{table*}

\begin{table*}[ht]
\centering
\caption{Ablation evaluation of CORSAIR in Scan2CAD \cite{Avetisyan_Scan2CAD_2019}. We compare our top-1 retrieved CAD (Top-1 Retrieval) and the ground-truth CAD annotation (GT annotation). We also compare our symmetry-aware matching method (w/ sym) with naive nearest neighbor (w/o sym). The percentage of test cases lower than different thresholds and the average single-direction Chamfer distance are reported.}
\label{table:pose_scan2cad}
\resizebox{\linewidth}{!}{\begin{tabular}{c|c|c|ccc|ccc|c}
\hline
Category & CAD model & Registration & RRE $\leq 5$º & RRE $\leq 15$º & RRE $\leq 45$º & RTE$\leq 0.05$ & RTE$\leq 0.10$ & RTE$\leq 0.15$ & $SCD(\times 10^{-2})$\\ \hline
\multirow{4}{*}{Chair} & \multirow{2}{*}{GT Annotation}    & w/o sym      
& 25.2    & 82.2     & 89.9 & 21.6 & 60.4 & 78.7 & 5.99      \\ 
\cline{3-3}
                       &                        & w/ sym       
& 34.4    & 87.9     & 94.0 & 27.4 & 68.5 & 85.2 & 5.43    \\ 
\cline{2-10}
                       & \multirow{2}{*}{Top-1 Retrieval} & w/o sym      
& 15.4    & 72.5     & 86.4 & 8.2 & 35.7 & 59.0 & 7.53        \\ 
\cline{3-3}
                       &                        & w/ sym       
& 23.2    & 78.5     & 88.9 & 9.7 & 40.5 & 63.5 & 6.81      \\ 
\cline{1-10}
\multirow{4}{*}{Table} & \multirow{2}{*}{GT Annotation}    & w/o sym      
& 19.2    & 58.4     & 72.5 & 12.7 & 38.5 & 58.8 & 7.20      \\ 
\cline{3-3}
                       &                        & w/ sym       
& 32.3    & 69.4     & 76.3 & 26.1 & 56.7 & 70.1 & 5.68      \\ 
\cline{2-10}
                       & \multirow{2}{*}{Top-1 Retrieval} & w/o sym      
& 11.7    & 40.9     & 52.9 & 4.8 & 17.5 & 28.5 & 9.06         \\ 
\cline{3-3}
                       &                        & w/ sym       
& 24.5    & 50.2     & 57.0 & 10.7 & 27.8 & 41.2 & 7.14 \\ 
\hline

\end{tabular}}
\end{table*}

The ShapeNet dataset \cite{Chang_shapenet_2015} contains a wide range of CAD models with category labels. We use point-clouds sampled from the CAD models provided by \cite{Yang_PointFlow_2019}. In the category of chair, we use 4612, 592, 1242 point-clouds for training, validation and testing. In the category of table, we use 5744, 778, 1557 point-clouds for training, validation and testing. The shape retrieval and pose estimation tasks are performed within each category. 
We first train the local feature extractor with positive and negative matching pairs as mentioned in eq. \eqref{eq:pairs} for 100 epochs. Then we freeze the parameters and train the embedding network with triplets $(\mathbf{Y}, \mathbf{P}, \mathbf{N})$ for another 100 epochs. Random rotation is applied to each of $\mathbf{Y}$, $\mathbf{P}$ and $\mathbf{N}$ before training and evaluating. The random seed is fixed in all the experiments for fair comparisons.

In ShapeNet, pose registration and the retrieval are considered as two separate tasks. To evaluate the pose registration performance, we estimate the transformation between a CAD model $\mathbf{Y}_i$ and a similar object in $\mathbf{Y}_j \in \mathcal{P_G}(\bfY_i)$ and measure the RRE and RTE. We assume that the symmetry of the $\mathbf{Y}_j$ is known and $\mathbf{Y}_i$ shares the same symmetry. In the evaluation of retrieval, we use the metric defined in \ref{sec:metric}. We report Precision@M$=0.1n$, where $n$ is the size of the test set. 

In the pose registration task, we compare our learned local feature (based on FCGF) with the hand-crafted FPFH feature \cite{Rusu_FPFH_2009}. FCGF is the same local feature extractor as ours but it generates matching pairs without our symmetry-aware method. RANSAC is applied to estimate the pose with given matching pairs in \eqref{eq:feature_pairs} with $K=5$. Fig.~\ref{fig:matching} is a qualitative result showing that our method generates more accurate matching pairs with the aid of symmetry information than the naive nearest-neighbor method. Mismatches caused by symmetry areas are filtered out. The quantitative results are shown in Table \ref{table:pose_shapenet}. Our symmetry-aware method outperforms FCGF baseline by 21.3\% and 29.1\% for chairs and tables, in terms of the ratio of test cases with RRE $\leq 5$º. The results show that our symmetry-aware method improves the pose estimation performance by refining matching pairs.


\begin{table}[t]
\centering
\caption{Retrieval quantitative results in ShapeNet \cite{Chang_shapenet_2015}.}
\label{table:retrieval_shapenet}
\scalebox{0.9}{

\begin{tabular}{c|cc|cc}
\multirow{2}{*}{Method} & \multicolumn{2}{c|}{Chair} & \multicolumn{2}{c}{Table} \\
                        & Precision@M   & Top-1 CD   & Precision@M   & Top-1 CD  \\ 
\hline
3D ResNet18 \cite{he2016deep}    &    21.81 & 0.182 & 17.49 & 0.231    \\
PointNet \cite{qi2017pointnet}  &    25.65    & 0.188 & 17.76 & 0.234          \\
FCGF \cite{Choy_FCGF_ICCV2019}  &    31.83    & 0.132 & 36.19 & 0.135          \\
CORSAIR (Ours)                            &    51.47    & 0.115 & 57.77 & 0.112          \\
\end{tabular}
}
\end{table}

We compare our retrieval module with other prevalent global shape descriptors including 3D ResNet18 \cite{he2016deep} and PointNet \cite{qi2017pointnet} as well as FCGF without our global embedding network. We use the 3D ResNet18 implementation in \cite{Choy_4D_2019}. Both 3D ResNet18, PointNet and our method generate a 256-D global descriptor for retrieval. The FCGF method directly uses latent vector $\mathbf{Z}(\mathbf{X})$ as the global feature, and the distance is measured by the Chamfer distance \eqref{eq:chamfer} in 256-D. All the methods are trained with the same loss function as defined in \eqref{eq: global triplet loss}. Quantitative results are shown in Table ~\ref{table:retrieval_shapenet}. Our method outperforms the baseline by a large margin in both Precision@M and Top-1 Chamfer distance metrics.


\subsection{Scan2CAD}

The Scan2CAD dataset \cite{Avetisyan_Scan2CAD_2019} provides object-level human-generated annotations. The annotations includes category label, segmentation, a similar CAD model in ShapeNet, and the corresponding pose. In this dataset, we assume that the segmentation and category are known but the annotated similar CAD model and the pose are unknown. We use the object segmentation labels to segment the object meshes from the scene and sample points on the surfaces to convert them to point-clouds. In the chair category, we use 2896, 343, 993 scanned point-clouds for training, validation and testing. In the table category, we use 1164, 150, 291 scanned point-clouds for training, validation and testing. Since the object distribution in the scenes is not uniform, we split the dataset by scenes instead of objects. Given pretrained parameters from ShapeNet, we train the local feature extractor and the embedding network on the Scan2CAD dataset separately for 100 epochs each. In the training phase, random rotations are applied to both scanned objects and CAD models. In the evaluation phase, we set the CAD models in the canonical pose.

For the pose registration task, we assume the number of symmetry classes for the CAD models are known in advance, and the scanned objects share the same symmetry with retrieved CAD models. For the retrieval task, the database $\mathcal{Y}$ contains CAD models from the Scan2CAD dataset and belong to the same category as the scanned object. The size of the CAD model database is 652 and 830 for the chair and table categories, respectively. Unlike in the ShapeNet experiments, for a scanned object $\mathbf{X}$, we only consider the ground-truth annotated CAD model as positive object $\mathbf{P}$. The negative object $\mathbf{N}$ is randomly sampled from the negative set $\mathcal{N_G}(\bfP)$ with respect to the annotated positive object $\mathbf{P}$, since the similarity between partially scanned objects and a CAD model is not well-defined. The retrieval task is evaluated jointly with the pose estimation task. We report the RRE and RTE as well as the single direction Chamfer distance to assess the overall alignment quality.

\begin{figure}[t]
  \includegraphics[width=\linewidth]{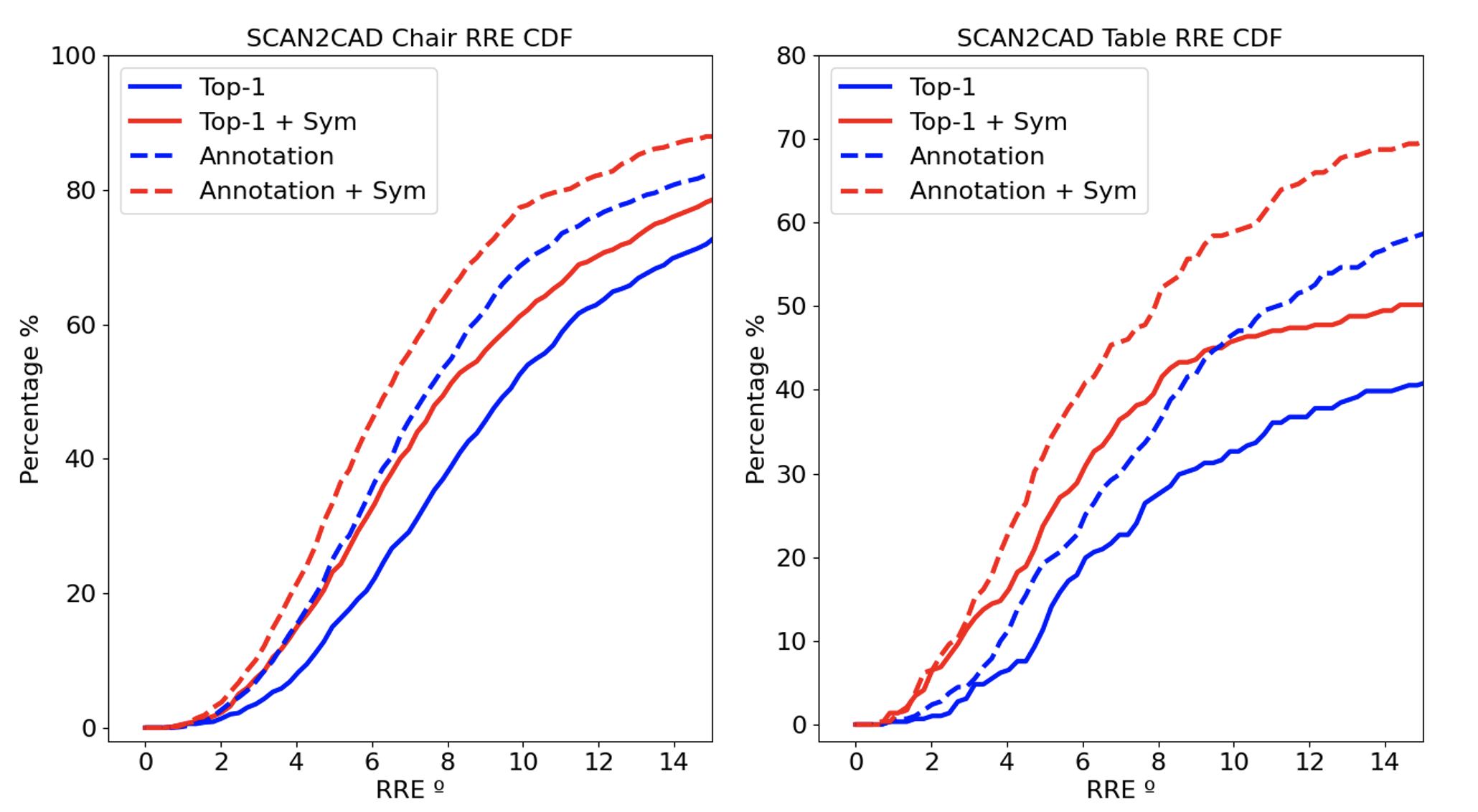}
  \caption{Cumulative distribution function of RRE on the Scan2CAD dataset \cite{Avetisyan_Scan2CAD_2019}. The solid lines represent RRE when aligning with top-1 retrieved CAD models. The dotted lines represents RRE aligning with annotated CAD models. Blue lines use naive nearest neighbor features + RANSAC for registration, while the red lines use our symmetry-aided nearest neighbor matching + RANSAC.}
  \label{fig:sca2cad-pose}
\end{figure}

We first evaluate the pose registration method by aligning scanned objects with the annotated CAD models. See Table \ref{table:pose_scan2cad}, Fig. \ref{fig:sca2cad-pose} for details. In a real world setting, our pose registration method is still able to estimate accurate poses (RRE $\leq$ $5^\circ$) for 25.2\% of the chairs and 19.2\% of the tables. The error for the majority of the test cases is within a reasonable range (RRE $\leq$ $15^\circ$). With our symmetry-aware method, we can further improve the ratio of accurate estimation by 9.2\% and 13.1\% for chairs and tables. The symmetry-aware method still generates better pose results in both categories when point-clouds are partially observed. Our method works well on approximately complete point-clouds, and for severely occluded scans we use the naive nearest neighbors with RANSAC as a back-up to handle asymmetric cases. 
Then we evaluate both the retrieval and the pose estimation using single direction Chamfer distance in the last column of Table \ref{table:pose_scan2cad}. Our retrieved CAD models can reach comparable alignment results compared with human-labeled ground-truth CAD models. This indicate that our method is able to retrieve reasonable models for better alignment.

\begin{figure*}[ht]
  \includegraphics[width=\linewidth,trim=0 10 0 10,clip]{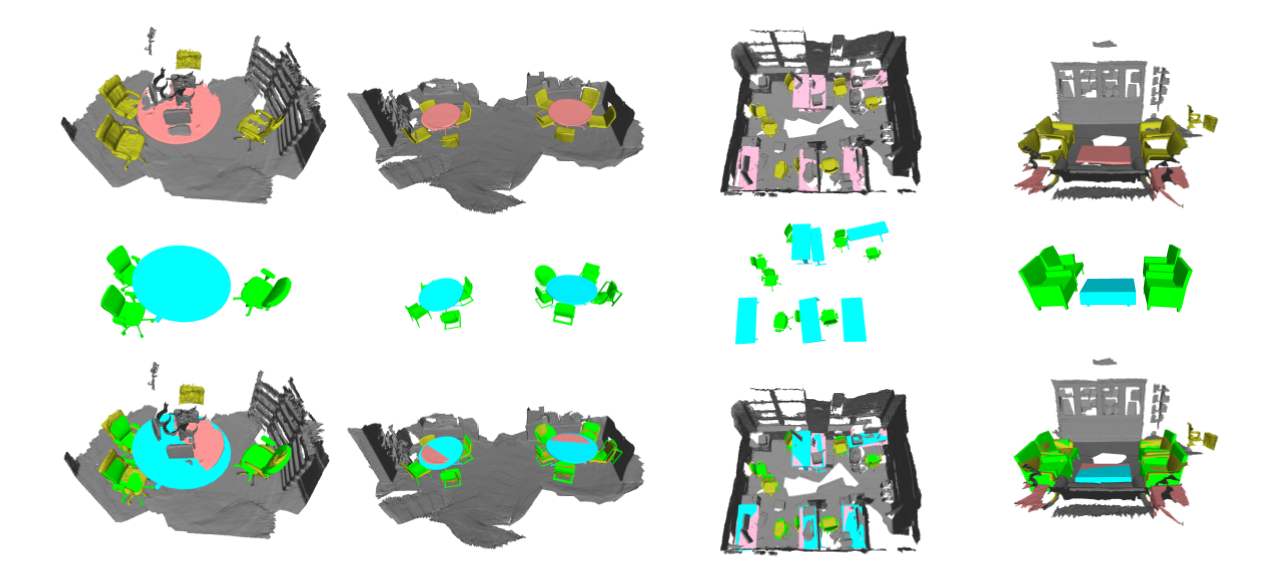}
  \caption{Scene-level reconstruction of scenes 0314\_00, 0355\_00, 0653\_00 and 0690\_00 from the Scan2CAD dataset \cite{Avetisyan_Scan2CAD_2019}. The segmented chairs (yellow) and tables (pink) in the first row are inputs to our model. The second row shows the predicted object map, obtained from aligning retrieved instances to the query point-clouds using CORSAIR. The retrieved chairs (green) and tables (blue) are overlaid back into the scene to visualize the reconstruction qualitatively.}
  \label{fig:scene_level}
\end{figure*}

In Fig. \ref{fig:scene_level}, we visualize the scene-level reconstruction to show qualitative results in real-world scenarios. Our method is able to retrieve CAD models that share the same structure with the scanned objects and align them accurately. Some failed cases are also presented, e.g., the chair on the right of the second scene. Most of the failed cases are caused by severe occlusions. The absence of key structures, like the legs of a table or chair, may lead to multiple solutions. The limitations of raw point-cloud measurements make it hard for our method to solve this kind of problems.

\section{Conclusion}
\label{sec:conclusion}

This work proposed CORSAIR, an approach for category-level retrieval and registration. CORSAIR is a fully convolutional model for point-cloud processing which jointly generates local point-wise geometric features and a global rotation-invariant shape feature. The global feature allows retrieval of similar object instances from the same category, while the local features, aided by symmetry class labeling, provide matching pairs for pose registration between the retrieved and query objects. For retrieval, CORSAIR outperforms the baseline methods by a large margin and even achieves comparable results when compared with human annotations. The symmetry-aware method proposed in CORSAIR refines the matching pair based on the naive nearest neighbor method and leads to considerable improvement on pose registration. Currently the global and local features extraction takes 0.2s for each object. While the retrieval step can run at 300 Hz, the registration using RANSAC runs slower than 1 Hz. We will explore faster alternatives for robust registration. Future work will also focus on making the pose estimation stage differentiable as well to enable end-to-end training of the whole model, using only pose annotations.



\balance
{\small
\bibliographystyle{cls/IEEEtran}
\bibliography{bib/ref.bib}
}
\end{document}